\documentclass{elsarticle}

\biboptions{authoryear}

\usepackage{graphicx} \graphicspath{{images/}}
\usepackage{tabularx}
\usepackage{multirow}

\usepackage{amssymb}
\usepackage{amsmath}

\usepackage{paralist}
\renewenvironment{itemize}{\begin{compactitem}}{\end{compactitem}}

\usepackage[textsize=scriptsize,textwidth=4cm]{todonotes}

\usepackage{hyperref}

\begin{document}

\title{NiftyNet: a deep-learning platform for medical imaging}

\author[weiss,cmic]{Eli  Gibson\fnref{jointfirst}}
\author[weiss]{Wenqi Li\fnref{jointfirst}\corref{cor1}}
\author[cmic]{Carole Sudre}
\author[weiss]{Lucas Fidon}
\author[weiss]{Dzhoshkun I. Shakir}
\author[weiss]{Guotai Wang}
\author[cmic]{Zach Eaton-Rosen}
\author[ion]{Robert Gray}
\author[weiss]{Tom Doel}
\author[cmic]{Yipeng Hu}
\author[cmic]{Tom Whyntie}
\author[ion]{Parashkev Nachev}
\author[cmic]{Marc Modat}
\author[weiss,cmic]{Dean C. Barratt}
\author[weiss]{S\'ebastien Ourselin}
\author[cmic]{M. Jorge Cardoso\fnref{jointlast}}
\author[weiss]{Tom Vercauteren\fnref{jointlast}}

\address[weiss]{Wellcome / EPSRC Centre for Interventional and Surgical
  Sciences (WEISS),\\ University College London, UK}

\address[cmic]{Centre
	for Medical Image Computing (CMIC), Departments of Medical Physics
	\& Biomedical Engineering and Computer Science, University College
	London, UK}

\address[ion]{Institute of Neurology, University College London, UK \& National Hospital for Neurology and Neurosurgery, London, UK}

\fntext[jointfirst]{Wenqi Li and Eli Gibson contributed equally to this work.}
\fntext[jointlast]{M. Jorge Cardoso and Tom Vercauteren contributed equally to this work.}

\cortext[cor1]{Corresponding author\\
  Email: wenqi.li@ucl.ac.uk\\
  Mailing Address:\\
  Wellcome / EPSRC Centre for Interventional and Surgical Sciences\\
  University College London\\
  Gower Street\\
  London, United Kingdom, WC1E 6BT}

\let\inlinecode\texttt

%%%
\begin{abstract}
\emph{Background and Objectives}
Medical image analysis and computer-assisted intervention problems are increasingly being addressed with deep-learning-based solutions. Established deep-learning platforms are flexible but do not provide specific functionality for medical image analysis and adapting them for this domain of application requires substantial implementation effort. Consequently, there has been substantial duplication of effort and incompatible infrastructure developed across many research groups. This work presents the open-source NiftyNet platform for deep learning in medical imaging. The ambition of NiftyNet is to accelerate and simplify the development of these solutions, and to provide a common mechanism for disseminating research outputs for the community to use, adapt and build upon.

\emph{Methods}
The NiftyNet infrastructure provides a modular deep-learning pipeline for a range of medical imaging applications including segmentation, regression, image generation and representation learning applications. Components of the NiftyNet pipeline including data loading, data augmentation, network architectures, loss functions and evaluation metrics are tailored to, and take advantage of, the idiosyncracies of medical image analysis and computer-assisted intervention. NiftyNet is built on the TensorFlow framework and supports features such as TensorBoard visualization of 2D and 3D images and computational graphs by default.

\emph{Results}
We present three illustrative medical image analysis applications built using NiftyNet infrastructure: (1) segmentation of multiple abdominal organs from computed tomography;  (2) image regression to predict computed tomography attenuation maps from brain magnetic resonance images; and (3) generation of simulated ultrasound images for specified anatomical poses.

\emph{Conclusions}
The NiftyNet infrastructure enables researchers to rapidly develop and distribute deep learning solutions for segmentation, regression, image generation and representation learning applications, or extend the platform to new applications.
\end{abstract}

%%%
\begin{keyword}
medical image analysis \sep deep learning \sep convolutional neural
network \sep segmentation \sep image regression \sep generative adversarial network
\end{keyword}

%%%
\maketitle

%%%
\section{Introduction}
Computer-aided analysis of medical images plays a critical role at many stages of the clinical workflow from population screening and diagnosis to treatment delivery and monitoring. This role is poised to grow as analysis methods become more accurate and cost effective. In recent years, a key driver of such improvements has been the adoption of deep learning and convolutional neural networks in many medical image analysis and computer-assisted intervention tasks.

Deep learning refers to a deeply nested composition of many simple functions (principally linear combinations such as convolutions, scalar non-linearities and moment normalizations) parameterized by variables. The particular composition of functions, called the architecture, defines a parametric function (typically with millions of parameters) that can be optimized to minimize an objective, or `loss', function, usually using some form of gradient descent.

Although the first use of neural networks for medical image analysis dates back more than twenty years~\citep{Lo:TMI:1995}, their usage has increased by orders of magnitude in the last five years. Recent reviews~\citep{Shen:ARBE:2017,Litjens:arXiv:2017} have highlighted that deep learning has been applied to a wide range of medical image analysis tasks (segmentation, classification, detection, registration, image reconstruction, enhancement, etc.) across a wide range of anatomical sites (brain, heart, lung, abdomen, breast, prostate, musculature, etc.). Although each of these applications have their own specificities, there is substantial overlap in software pipelines implemented by many research groups.

Deep-learning pipelines for medical image analysis comprise many interconnected components. Many of these are common to all deep-learning pipelines:
\begin{itemize}
	\item separation of data into training, testing and validation sets;
	\item randomized sampling during training;
	\item image data loading and sampling;
	\item data augmentation;
	\item a network architecture defined as the composition of many simple functions;
	\item a fast computational framework for optimization and inference;
	\item metrics for evaluating performance during training and inference.
\end{itemize}

In medical image analysis, many of these components have domain specific idiosyncrasies, detailed in Section~\ref{sec:medical_domain}. For example, medical images are typically stored in specialized formats that handle large 3D images with anisotropic voxels and encode additional spatial information and/or patient information, requiring different data loading pipelines. Processing large volumetric images has high memory requirements and motivates domain-specific memory-efficient networks or custom data sampling strategies. Images are often acquired in standard anatomical views and can represent physical properties quantitatively, motivating domain-specific data augmentation and model priors. Additionally, the clinical implications of certain errors may warrant custom evaluation metrics. Independent reimplementation of all of this custom infrastructure results in substantial duplication of effort, poses a barrier to dissemination of research tools and inhibits fair comparisons between competing methods.

This work presents the open-source NiftyNet\footnote{Available at \url{http://niftynet.io}}
platform to 1) facilitate efficient deep learning research in medical image analysis and computer-assisted intervention; and 2) reduce duplication of effort. The NiftyNet platform comprises an implementation of the common infrastructure and common networks used in medical imaging, a database of pre-trained networks for specific applications and tools to facilitate the adaptation of deep learning research to new clinical applications with a shallow learning curve.

%%%
\section{Background}
The development of common software infrastructure for medical image analysis and computer-assisted intervention has a long history. Early efforts included the development of medical imaging file formats (e.g. ACR-NEMA (1985), Analyze 7.5 (1986), DICOM (1992) MINC (1992), and NIfTI (2001)). Toolsets to solve common challenges such as registration (e.g. NiftyReg~\citep{Modat:CMPB:2010}, ANTs~\citep{Avants:NEUROIM:2011} and elastix~\citep{Klein:TMI:2010}), segmentation (e.g. NiftySeg~\citep{Cardoso:ISBI:2012}), and biomechanical modeling (e.g.~\citep{Johnsen:IJCARS:2015}) are available for use as part of image analysis pipelines. Pipelines for specific research applications such as FSL~\citep{Smith:NEUROIM:2004} for functional MRI analysis and Freesurfer~\citep{Fischl:NEUROIM:1999,Dale:NEUROIM:1999} for structural neuroimaging have reached widespread use. More general toolkits offering standardized implementations of algorithms (VTK and ITK~\citep{Pieper:ISBI:2006}) and application frameworks (NifTK~\citep{Clarkson:IJCARS:2015}, MITK~\citep{Nolden:IJCARS:2013} and 3D Slicer~\citep{Pieper:ISBI:2006}) enable others to build their own pipelines. Common software infrastructure has supported and accelerated medical image analysis and computer-assisted intervention research across hundreds of research groups. However, despite the wide availability of general purpose deep learning software tools, deep learning technology has limited support in current software infrastructure for medical image analysis and computer-assisted intervention.

Software infrastructure for general purpose deep learning is a recent development. Due to the high computational demands of training deep learning models and the complexity of efficiently using modern hardware resources (general purpose graphics processing units and distributed computing, in particular), numerous deep learning libraries and platforms have been developed and widely adopted, including cuDNN~\citep{Chetlur:arXiv:2014}, TensorFlow~\citep{Abadi:arXiv:2016}, Theano ~\citep{Bastien:DeepLearning:2012}, Caffe~\citep{Jia:ACMMM:2014}, Torch~\citep{Collobert:BigLearn:2011}, CNTK~\citep{Seide:KDD:2016}, and MatConvNet~\citep{Vedaldi:ACMM:2015}.

These platforms facilitate the definition of complex deep learning networks as compositions of simple functions, hide the complexities of differentiating the objective function with respect to trainable parameters during training, and execute efficient implementations of performance-critical functions during training and inference. These frameworks have been optimized for performance and flexibility, and using them directly can be challenging, inspiring the development of platforms that simplify the development process for common usage scenarios, such as Keras~\citep{Chollet:github:2015}, and TensorLayer~\citep{Dong:ACMM:2017} for TensorFlow and Lasagne~\citep{Dielman:Zenodo:2015} for Theano.
However, by avoiding assumptions about the application to remain general, the platforms are unable to provide specific functionality for medical image analysis and adapting them for this domain of application requires substantial implementation effort.

Developed concurrently with the NiftyNet platform, the Deep Learning Toolkit\footnote{\url{https://dltk.github.io}} aims to support fast prototyping and reproducibility by implementing deep learning methods and modules for medical image analysis. While still in preliminary development, it appears to focus on deep learning building blocks rather than analysis pipelines. NifTK~\citep{Clarkson:IJCARS:2015,Gibson:SPIE:2017} and Slicer3D (via the DeepInfer~\citep{Mehrtash:SPIE:2017} plugin) provide infrastructure for distribution of trained deep learning pipelines. Although this does not address the substantial infrastructure needed for training deep learning pipelines, integration with existing medical image analysis infrastructure and modular design makes these platforms promising routes for distributing deep-learning pipelines.
%%%
\section{Typical deep learning pipeline}
\begin{figure*}[htp]
  \centering
  \includegraphics[height=14cm]{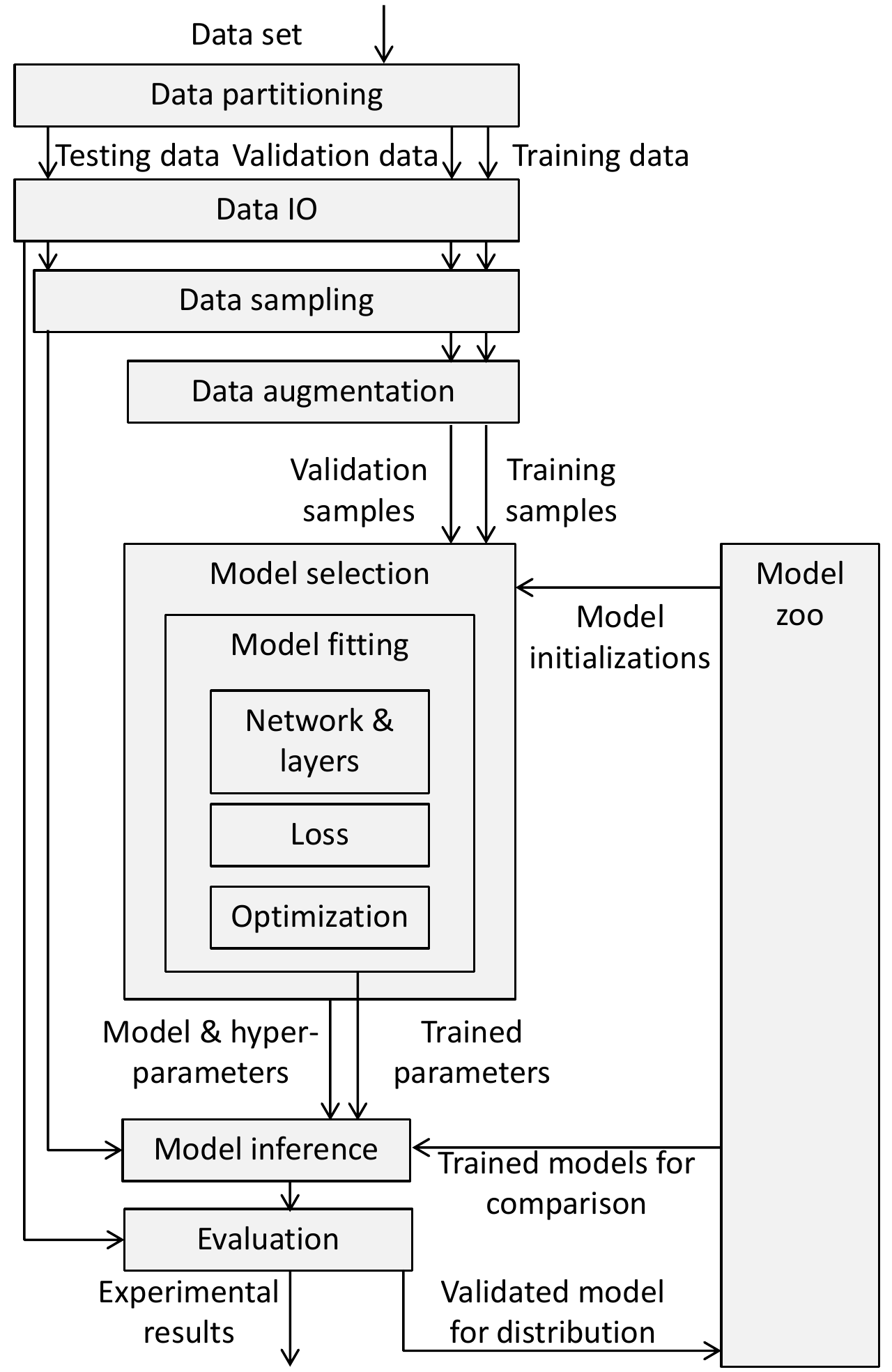}
  \caption{Data flow implemented in typical deep learning projects. Boxes represent the software infrastructure to be developed and arrows represent the data flow.}
  \label{fig:pipeline}
\end{figure*}
Deep learning adopts the typical machine learning pipeline consisting of three phases: model selection (picking and fitting a model on training data), model evaluation (measuring the model performance on testing data), and model distribution (sharing the model for use on a wider population). Within these simple phases lies substantial complexity, illustrated in Figure~\ref{fig:pipeline}. The most obvious complexity is in implementing the network being studied. Deep neural networks generally use simple functions, but compose them in complex hierarchies; researchers must implement the network being tested, as well as previous networks (often incompletely specified) for comparison. To train, evaluate and distribute these networks, however, requires further infrastructure. Data sets must be correctly partitioned to avoid biassed evaluations, sometimes considering data correlations (e.g. images acquired at the same hospital may be more similar to each other than to those from other hospitals). The data must be sampled, loaded and passed to the network, in different ways depending on the phase of the pipeline. Algorithms for tuning hyper-parameters within a family of models and optimizing model parameters on the training data are needed. Logging and visualization are needed to debug and dissect models during and after training. In applications with limited data, data sets must be augmented by perturbing the training data in realistic ways to prevent over-fitting. In deep learning, it is common practice to adapt previous network architectures, trained or untrained, in part or in full for similar or different tasks; this requires a community repository (popularly called a \emph{model zoo}) storing models and parameters in an adaptable format. Much of this infrastructure is recreated by each researcher or research group undertaking a deep learning project, and much of it depends on the application domain being addressed.

%%%
\section{Design considerations for deep learning in medical imaging}
\label{sec:medical_domain}
Medical image analysis differs from other domains where deep learning is applied due to characteristics of the data itself, and the applications in which they are used. In this section, we present the domain-specific requirements driving the design of NiftyNet.

%%%
\subsection{Data availability}
Acquiring, annotating and distributing medical image data sets have higher costs than in many computer vision tasks. For many medical imaging modalities, generating an image is costly. Annotating images for many applications requires high levels of expertise from clinicians with limited time. Additionally, due to privacy concerns, sharing data sets between institutions, let alone internationally, is logistically and legally challenging. Although recent tools such as DeepIGeoS~\citep{Wang:arXiv:2017} for semi-automated annotation and GIFT-Cloud~\citep{Doel:CMPB:2017} for data sharing are beginning to reduce these barriers, typical data sets remain small.  Using smaller data sets increases the importance of data augmentation, regularization, and cross-validation to prevent over-fitting. The additional cost of data set annotation also places a greater emphasis on semi- and unsupervised learning.

%%%
\subsection{Data dimensionality and size}
Data dimensionality encountered in medical image analysis and computer-assisted intervention typically ranges from 2D to 5D. Many medical images, including MRI, CT, PET and SPECT, capture volumetric images. Longitudinal imaging (multiple images taken over time) is typical in interventional settings as well as clinically useful for measuring organ function (e.g. blood ejection fraction in cardiac imaging) and disease progression (e.g. cortical thinning in neurodegenerative diseases).

At the same time, capturing high-resolution data in multiple dimensions is often necessary to detect small but clinically important anatomy and pathology. The combination of these factors results in large data sizes for each sample, which impact computational and memory costs. Deep learning in medical imaging uses various strategies to account for this challenge. Many networks are designed to use partial images: 2D slices sampled along one axis from 3D images~\citep{Zhou:LABELS:2016},
3D subvolumes~\citep{Li:IPMI:2017}, anisotropic convolution~\cite{Wang:arXiv:2017a}, or combinations of subvolumes along multiple axes~\citep{Roth:MICCAI:2014}. Other networks use multi-scale representations allowing deeper and wider networks on lower-resolution representations~\citep{Milletari:P3DV:2016,Kamnitsas:MEDIA:2017}. A third approach uses dense networks to reuse feature representations multiple times in the network~\citep{Gibson:MICCAI:2017}. Smaller batch sizes can reduce the memory cost, but rely on different weight normalization functions such as batch renormalization~\citep{Ioffe:arXiv:2017}, weight normalization~\citep{Salimans:arXiv:2016} or layer normalization~\citep{Ba:arXiv:2016}.
%%%
\subsection{Data formatting}
Data sets in medical imaging are typically stored in different formats than in many computer vision tasks. To support the higher-dimensional medical image data, specialized formats have been adopted (e.g. DICOM, NIfTI, Analyze). These formats frequently also store metadata that is critical to image interpretation, including spatial information (anatomical orientation and voxel anisotropy), patient information (demographics and identifiers), and acquisition information (modality types and scanner parameters). These medical imaging specific data formats are typically not supported by existing deep learning frameworks, requiring custom infrastructure for loading images.
%%%
\subsection{Data properties}
The characteristic properties of medical image content pose opportunities and challenges. Medical images are obtained under controlled conditions, allowing more predictable data distributions. In many modalities, images are calibrated such that spatial relationships and image intensities map directly to physical quantities and are inherently normalized across subjects. For a given clinical workflow, image content is typically consistent, potentially enabling the characterization of plausible intensity and spatial variation for data augmentation. However, some clinical applications introduce additional challenges. Because small image features can have large clinical importance, and because some pathology is very rare but life-threatening, medical image analysis must deal with large class imbalances, motivating special loss functions~\citep{Milletari:P3DV:2016,Fidon:arXiv:2017,Sudre:DLMIA:2017}. Furthermore, different types of error may have very different clinical impacts, motivating specialized loss functions and evaluation metrics (e.g. spatially weighted segmentation metrics). Applications in computer-assisted intervention where analysis results are used in real time (e.g. ~\cite{Gibson:SPIE:2017,Peraza-Herrera:IROS:2017}) have additional constraints on analysis latency.
%%%
\section{NiftyNet: a platform for deep learning in medical imaging}
The NiftyNet platform aims to augment the current deep learning infrastructure to address the ideosyncracies of medical imaging described in Section~\ref{sec:medical_domain}, and lower the barrier to adopting this technology in medical imaging applications. NiftyNet is built using the TensorFlow library, which provides the tools for defining computational pipelines and executing them efficiently on hardware resources, but does not provide any specific functionality for processing medical images, or high-level interfaces for common medical image analysis tasks. NiftyNet provides a high-level deep learning pipeline with components optimized for medical imaging applications (data loading, sampling and augmentation, networks, loss functions, evaluations, and a model zoo) and specific interfaces for medical image segmentation, classification, regression, image generation and representation learning applications.

%%%
\subsection{Design goals}
The design of NiftyNet follows several core principles which support a set of key requirements:
\begin{itemize}
	\item support a wide variety of application types in medical image analysis and computer-assisted intervention;
	\item enable research in one aspect of the deep learning pipeline without the need for recreating the other parts;
	\item be simple to use for common use cases, but flexible enough for complex use cases;
	\item support built-in TensorFlow features (parallel processing, visualization) by default;
	\item support best practices (data augmentation, data set separation) by default;
	\item support model distribution and adaptation.
\end{itemize}

%%%
\subsection{System overview}
\begin{figure*}[htp]
  \centering
  \includegraphics[height=9cm]{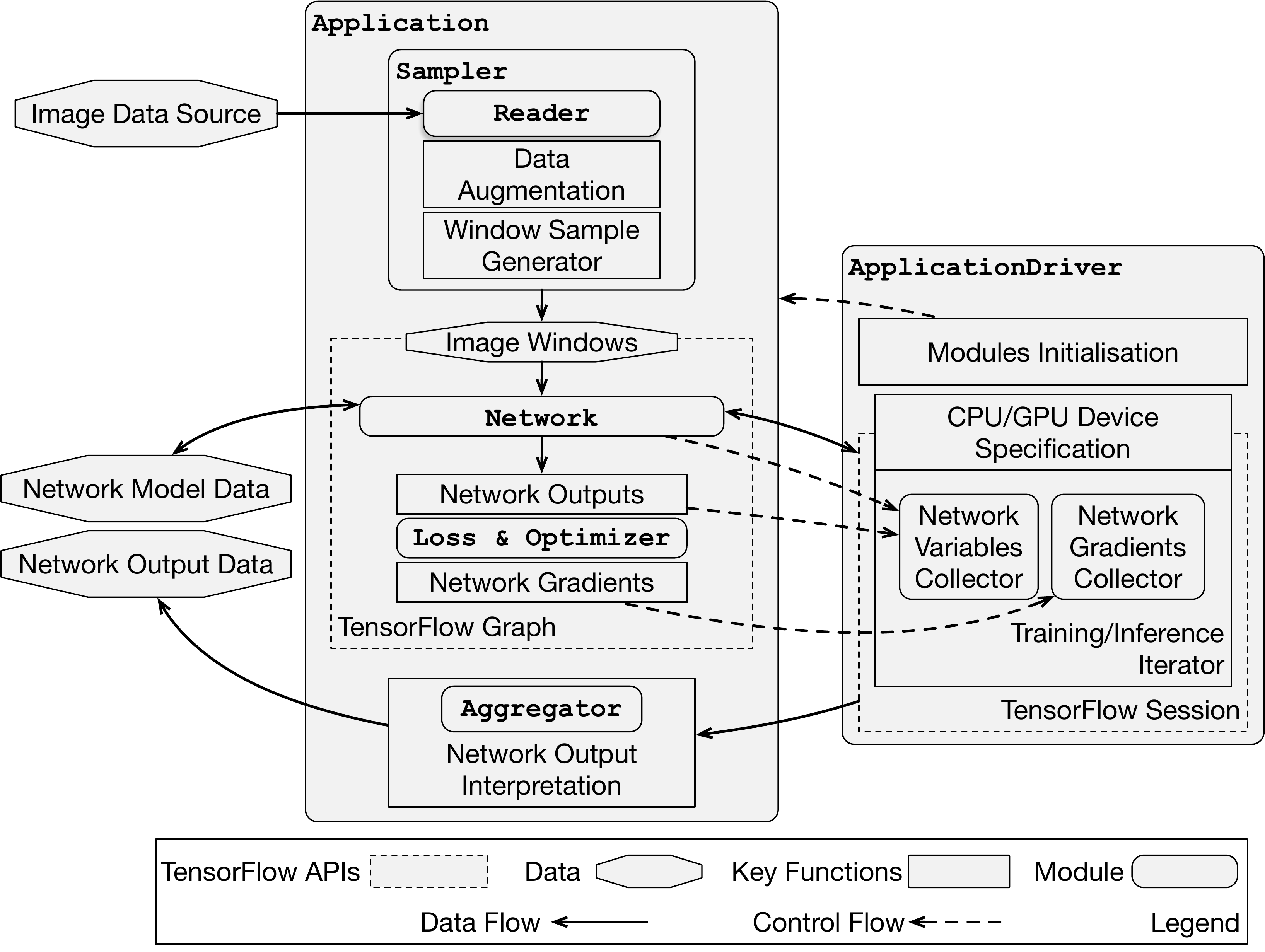}
  \caption{A brief overview of NiftyNet components.}
  \label{fig:diagram}
\end{figure*}
The NiftyNet platform comprises several modular components. The NiftyNet \inlinecode{ApplicationDriver} defines the common structure across all applications, and is responsible for instantiating the data analysis pipeline and distributing the computation across the available computational resources. The NiftyNet \inlinecode{Application} classes encapsulate standard analysis pipelines for different medical image analysis applications, by connecting four components: a \inlinecode{Reader} to load data from files, a \inlinecode{Sampler} to generate appropriate samples for processing, a \inlinecode{Network} to process the inputs, and an output handler (comprising the \inlinecode{Loss} and \inlinecode{Optimizer} during training and an \inlinecode{Aggregator} during inference and evaluation).  The \inlinecode{Sampler} includes sub-components for data augmentation. The \inlinecode{Network} includes sub-components representing individual network blocks or larger conceptual units.  These components are briefly depicted in Figure~\ref{fig:diagram} and detailed in the following sections.

As a concrete illustration, one instantiation of the \inlinecode{SegmentationApplication} could use the following modules. During training, it could use a \inlinecode{UniformSampler} to generate small image patches and corresponding labels; a \inlinecode{vnet} \inlinecode{Network} would process batches of images to generate segmentations; a \inlinecode{Dice} \inlinecode{LossFunction} would compute the loss used for backpropagation using the \inlinecode{Adam} \inlinecode{Optimizer}.  During inference, it could use a \inlinecode{GridSampler} to generate a set of non-overlapping patches to cover the image to segment, the same network to generate corresponding segmentations, and a \inlinecode{GridSamplesAggregator} to aggregate the patches into a final segmentation.
%%%
\subsection{Component details: ApplicationDriver class}
The NiftyNet \inlinecode{ApplicationDriver} defines the common structure for all NiftyNet pipelines. It is responsible for instantiating the data and \inlinecode{Application} objects and distributing the workload across and recombining results from the computational resources (potentially including multiple CPUs and GPUs). It is also responsible for handling variable initialization, variable saving and restoring, and logging. Implemented as a template design pattern~\citep{Gamma:1995}, the \inlinecode{ApplicationDriver} delegates application-specific functionality to separate \inlinecode{Application} classes.

The \inlinecode{ApplicationDriver} can be configured from the command line or programmatically using a human-readable configuration file. This file contains the data set definitions and all the settings that deviate from the defaults. When the \inlinecode{ApplicationDriver} saves its progress, the full configuration (including default parameters) is also saved so that the analysis pipeline can be recreated to continue training or carry out inference internally or with a distributed model.
%%%
\subsection{Component details: Application class}
Medical image analysis encompasses a wide range of tasks for different parts of the pre-clinical and clinical workflow: segmentation, classification, detection, registration, reconstruction, enhancement, model representation and generation. Different applications use different types of inputs and outputs, different networks, and different evaluation metrics; however, there is common structure and functionality among these applications supported by NiftyNet. NiftyNet currently supports
\begin{itemize}
\item image segmentation,
\item image regression,
\item image model representation (via auto-encoder applications), and
\item image generation (via auto-encoder and generative adversarial networks (GANs)),
\end{itemize}
and it is designed in a modular way to support the addition of new application types, by encapsulating typical application workflows in \inlinecode{Application} classes.

The \inlinecode{Application} class defines the required data interface for the \inlinecode{Network} and \inlinecode{Loss}, facilitates the instantiation of appropriate \inlinecode{Sampler} and output handler objects, connects them as needed for the application, and specifies the training regimen. For example, the \inlinecode{SegmentationApplication} specifies that networks accept images (or patches thereof) and generate corresponding labels, that losses accept generated and reference segmentations and an optional weight map, and that the optimizer trains all trainable variables in each iteration. In contrast, the \inlinecode{GANApplication} specifies that networks accept a noise source, samples of real data and an optional conditioning image, losses accept logits denoting if a sample is real or generated, and the optimizer alternates between training the discriminator sub-network and the generator sub-network.
%%%
\subsection{Component details: Networks and Layers}
\begin{figure}[htp]
	\centering
	\includegraphics[height=0.88\textheight]{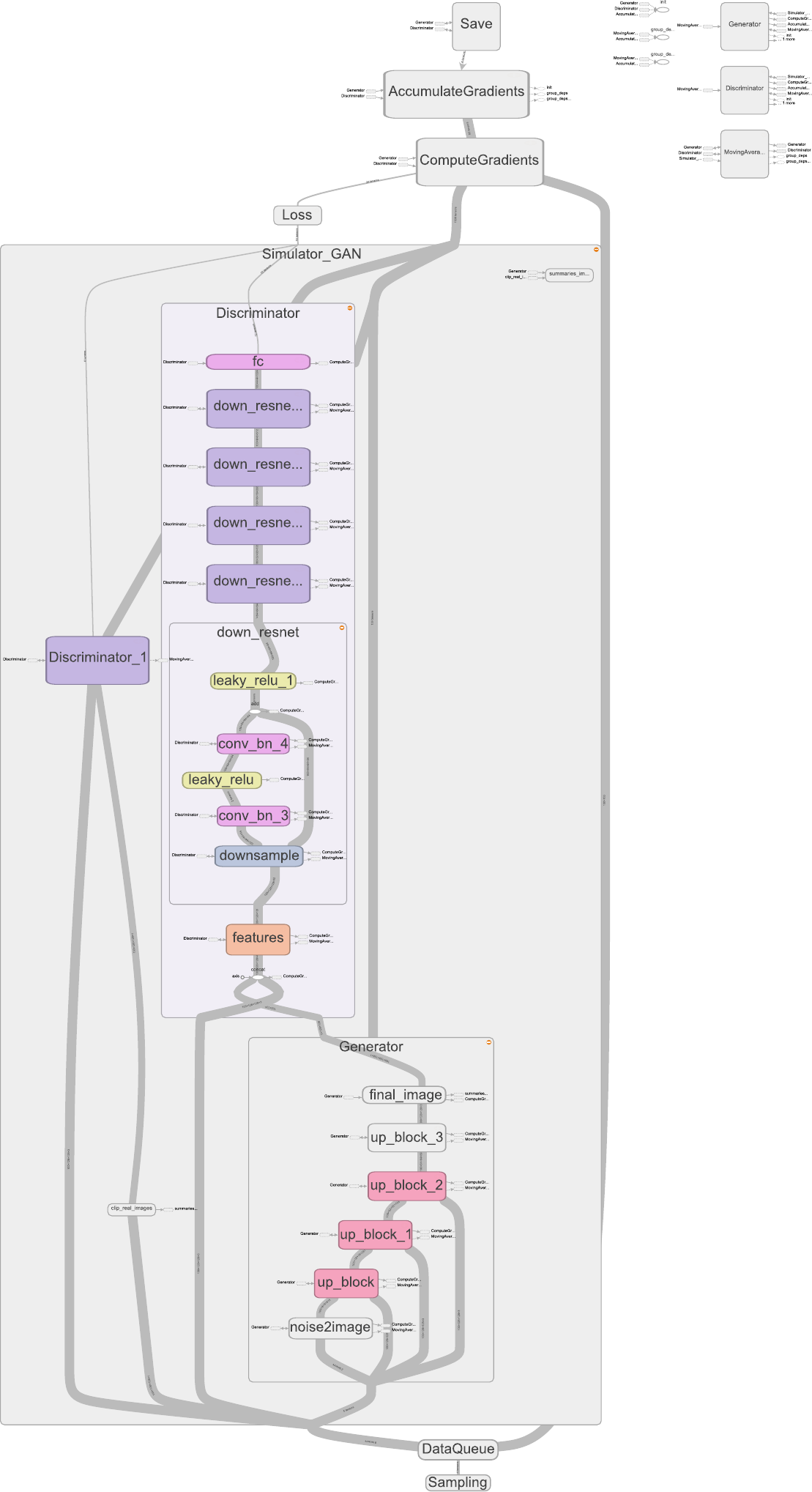}
	\caption{TensorBoard visualization of a NiftyNet generative adversarial network. TensorBoard interactively shows the composition of conceptual blocks (rounded rectangles) and their interconnections (grey lines) and color-codes similar blocks. Above, the generator and discriminator blocks and one of the discriminator's residual blocks are expanded. Font and block sizes were edited for readability.}
	\label{fig:gan_tensorboard}
\end{figure}

The complex composition of simple functions that comprise a deep learning architecture is simplified in typical networks by the repeated reuse of conceptual blocks. In NiftyNet, these conceptual blocks are represented by encapsulated \inlinecode{Layer} classes, or inline using TensorFlow's scoping system. Composite layers, and even entire networks, can be constructed as simple compositions of NiftyNet layers and TensorFlow operations. This supports the reuse of existing networks by clearly demarcating conceptual blocks of code that can be reused and assigning names to corresponding sets of variables that can be reused in other networks (detailed in Section~\ref{sec:modelzoo}). This also enables automatic support for visualization of the network graph as a hierarchy at different levels of detail using the TensorBoard visualizer~\citep{Mane:github:2015} as shown in Figure~\ref{fig:gan_tensorboard}. Following the model used in Sonnet~\citep{Reynolds:github:2017}, \inlinecode{Layer} objects define a scope upon instantiation, which can be reused repeatedly to allow complex weight-sharing without breaking encapsulation.
%%%
\subsection{Component details: data loading}
The \inlinecode{Reader} class is responsible for loading corresponding image files from medical file formats for a specified data set, and applying image-wide preprocessing. For simple use cases, NiftyNet can automatically identify corresponding images in a data set by searching a specified file path and matching user-specified patterns in file names, but it also allows explicitly tabulated comma-separated value files for more complex data set structures (e.g. cross-validation studies). Input and output of medical file formats are already supported in multiple existing Python libraries, although each library supports different sets of formats. To facilitate a wide range of formats, NiftyNet uses \inlinecode{nibabel}~\citep{Brett:Zenodo:2016} as a core dependency but can fall back on other libraries (e.g. SimpleITK~\citep{Lowekamp:FRONTNEUROINFORM:2013} if they are installed and a file format is not supported by \inlinecode{nibabel}. A pipeline of image-wide preprocessing functions, described in Section~\ref{sec:preprocessing}, is applied to each image before samples are taken.
%%%
\subsection{Component details: Samplers and output handlers}
To handle the breadth of applications in medical image analysis and computer-assisted intervention, NiftyNet provides flexibility in mapping from an input data set into packets of data to be processed and from the processed data into useful outputs. The former is encapsulated in \inlinecode{Sampler} classes, and the latter is encapsulated in output handlers. Because the sampling and output handling are tightly coupled and depend on the action being performed (i.e. training, inference or evaluation), the instantiation of matching \inlinecode{Sampler} objects and output handlers is delegated to the \inlinecode{Application} class.

\inlinecode{Sampler} objects generate a sequence of packets of corresponding data for processing. Each packet contains all the data for one independent computation (e.g. one step of gradient descent during training), including images, labels, classifications, noise samples or other data needed for processing. During training, samples are taken randomly from the training data, while during inference and evaluation the samples are taken systematically to process the whole data set. To feed these samples to TensorFlow, NiftyNet automatically takes advantage of TensorFlow's data queue support: data can be loaded and sampled in multiple CPU threads, combined into mini-batches and consumed by one or more GPUs. NiftyNet includes \inlinecode{Sampler} classes for sampling image patches (uniformly or based on specified criteria), sampling whole images rescaled to a fixed size and sampling noise; and it supports composing multiple \inlinecode{Sampler} objects for more complex inputs.

Output handlers take different forms during training and inference. During training, the output handler takes the network output, computes a loss and the gradient of the loss with respect to the trainable variables, and uses an inlinecode{Optimizer} to iteratively train the model. During inference, the output handler generates useful outputs by aggregating one or more network outputs and performing any necessary postprocessing (e.g. resizing the outputs to the original image size). NiftyNet currently supports \inlinecode{Aggregator} objects for combining image patches, resizing images, and computing evaluation metrics.
%%%
\subsection{Component details: data normalization and augmentation}
\label{sec:preprocessing}
Data normalization and augmentation are two approaches to compensating for small training data sets in medical image analysis, wherein the training data set is too sparse to represent the variability in the distribution of images. Data normalization reduces the variability in the data set by transforming inputs to have specified invariant properties, such as fixed intensity histograms or moments (mean and variance). Data augmentation artificially increases the variability of the training data set by introducing random perturbations during training, for example applying random spatial transformations or adding random image noise. In NiftyNet, data augmentation and normalization are implemented as \inlinecode{Layer} classes applied in the \inlinecode{Sampler}, as plausible data transformations will vary between applications. Some of these layers, such as histogram normalization, are data dependent; these layers compute parameters over the data set before training begins. NiftyNet currently supports mean, variance and histogram intensity data normalization, and flip, rotation and scaling spatial data augmentation.
%%%
\subsection{Component details: data evaluation}
Summarizing and comparing the performance of image analysis pipelines typically rely on standardized descriptive metrics and error metrics as surrogates for performance. Because individual metrics are sensitive to different aspects of performance, multiple metrics are reported together. Reference implementations of these metrics reduce the burden of implementation and prevent implementation inconsistencies. NiftyNet currently supports the calculation of descriptive and error metrics for segmentation. Descriptive statistics include spatial metrics (e.g. volume, surface/volume ratio, compactness) and intensity metrics (e.g. mean, quartiles, skewness of intensity). Error metrics, computed with respect to a reference segmentation, include overlap metrics (e.g. Dice and Jaccard scores; voxel-wise sensitivity, specificity and accuracy), boundary distances (e.g. mean absolute distance and Hausdorff distances) and region-wise errors (e.g. detection rate; region-wise sensitivity, specificity and accuracy).
%%%
\subsection{Component details: model zoo for network reusability}
\label{sec:modelzoo}
To support the reuse of network architectures and trained models, many deep learning platforms host a database of existing trained and untrained networks in a standardized format, called a model zoo. Trained networks can be used directly (as part of a workflow or for performance comparisons), fine-tuned for different data distributions (e.g. a  different hospital's images), or used to initialize networks for other applications (i.e. transfer learning). Untrained networks or conceptual blocks can be used within new networks. NiftyNet provides several mechanisms to support the distribution and reuse of networks and conceptual blocks.

Trained NiftyNet networks can be restored directly using configuration options. Trained networks developed outside of NiftyNet can be adapted to NiftyNet by encapsulating the network within a \inlinecode{Network} class derived from \inlinecode{TrainableLayer}. Externally trained weights can be loaded within NiftyNet using a \inlinecode{restore$\_$initializer}, adapted from Sonnet~\citep{Reynolds:github:2017}, for the complete network or individual conceptual blocks. \inlinecode{restore$\_$initializer} initializes the network weights with those stored in a specified checkpoint, and supports \inlinecode{variable$\_$scope} renaming for checkpoints with incompatible scope names. Smaller conceptual blocks, encapsulated in \inlinecode{Layer} classes, can be reused in the same way. Trained networks incorporating previous networks are saved in a self-contained form to minimize dependencies.

The NiftyNet model zoo contains both untrained networks (e.g. \inlinecode{unet}~\citep{Cicek:MICCAI:2016} and \inlinecode{vnet}~\citep{Milletari:P3DV:2016} for segmentation), as well as trained networks for some tasks (e.g. \inlinecode{dense$\_$vnet}~\citep{Gibson:TMI:2017} for multi-organ abdominal CT segmentation, \inlinecode{wnet}~\citep{Wang:arXiv:2017a} for brain tumor segmentation and \inlinecode{simulator$\_$gan}~\citep{Hu:RAMBO:2017} for generating ultrasound images). Model zoo entries should follow a standard format comprising:
\begin{itemize}
	\item Python source code defining any components not included in NiftyNet (e.g. external \inlinecode{Network} classes, \inlinecode{Loss} functions);
	\item an example configuration file defining the default settings and the data ordering;
	\item documentation describing the network and assumptions on the input data (e.g. dimensionality, shape constraints, intensity statistic assumptions).
\end{itemize}%
For trained networks, it should also include:
\begin{itemize}
	\item a Tensorflow checkpoint containing the trained weights;
	\item documentation describing the data used to train the network and on which the trained network is expected to perform adequately.
\end{itemize}
%%%
\subsection{Platform processes}
In addition to the implementation of common functionality, NiftyNet development has adopted good software development processes to support the ease-of-use, robustness and longevity of the platform as well as the creation of a vibrant community. The platform supports easy installation via the \inlinecode{pip} installation tool\footnote{\url{https://pip.pypa.io}} (i.e. \inlinecode{pip install niftynet}) and provides analysis pipelines that can be run as part of the command line interface. Examples demonstrating the platform in multiple use cases are included to reduce the learning curve. The NiftyNet repository uses continuous integration incorporating system and unit tests for regression testing. NiftyNet releases will follow the semantic versioning 2.0 standard~\citep{Preston-Werner:2015} to ensure clear communication regarding backwards compatibility.
%%%
\section{Results: illustrative applications}
%%%
\subsection{Abdominal organ segmentation}
Segmentations of anatomy and pathology on medical images can support image-guided interventional workflows by enabling the visualization of hidden anatomy and pathology during surgical navigation. Here we present an example, based on a simplified version of~\citep{Gibson:TMI:2017}, that illustrates the use of NiftyNet to train a Dense V-network segmentation network to segment organs on abdominal CT that are important to pancreatobiliary interventions: the gastrointestinal tract (esophagus, stomach and duodenum), the pancreas, and anatomical landmark organs (liver, left kidney, spleen and stomach).

The data used to train the network comprised 90 abdominal CT with manual segmentations from two publicly available data sets~\citep{Landman:BTCV:2015,Roth:TCIA:2015}, with additional manual segmentations performed at our centre.

\begin{figure}[!t]
  \centering
  \newlength{\imagewidth}
  \settowidth{\imagewidth}{\includegraphics{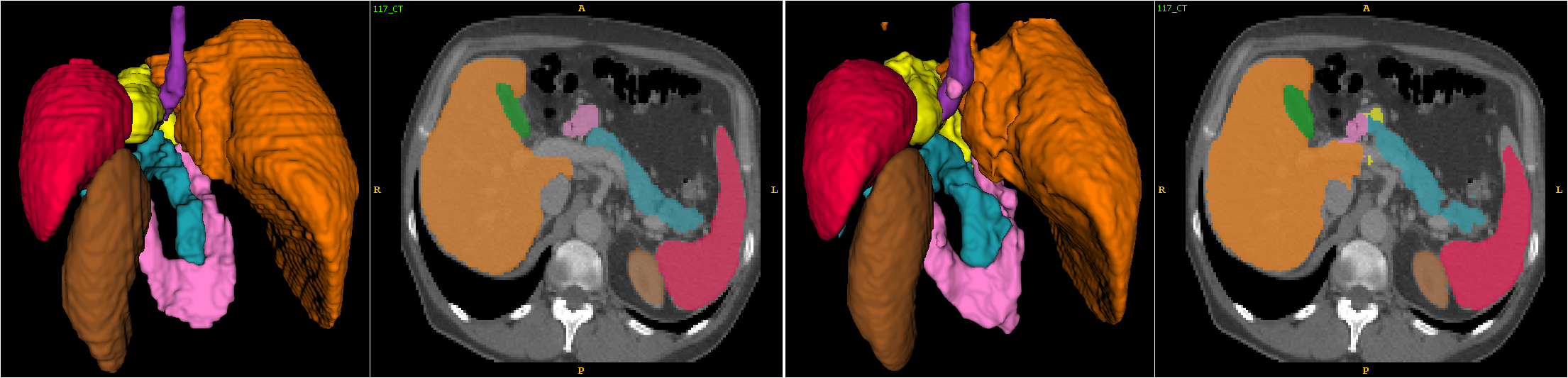}}
  \begin{tabular}{cc}%
\includegraphics[trim=0 0 0.765\imagewidth{} 0, clip, width=1.5in]{segmentation.png}&\includegraphics[trim=0.5\imagewidth{} 0 .265\imagewidth{} 0 0, clip, width=1.5in]{segmentation.png}\\
\includegraphics[trim=0.240\imagewidth{} 0 0.5\imagewidth{} 0, clip, width=1.5in]{segmentation.png}&\includegraphics[trim=0.740\imagewidth{} 0 0 0, clip, width=1.5in]{segmentation.png}\\
  	Reference standard & NiftyNet segmentation
  	\end{tabular}
  \caption{Reference standard (left) and NiftyNet (right) multi-organ abdominal CT segmentation for the subject with Dice scores closest to the median. Each segmentation is shown with a surface rendering view from the posterior direction and with organ labels overlaid on a transverse CT slice.}
  \label{fig:segmentation}
\end{figure}

\begin{table}[!b]
	\centering
	\caption{Median segmentation metrics for 8 organs aggregated over the 9-fold cross-validation.}
  \begin{tabular}{ccccc}
  	\hline\hline
  	& Dice  & Relative & Mean  & 95th Percentile \\
  	&  score &  Volume & Absolute  &  Hausdorff \\
  	&   &  Difference & Distance  &  Distance\\
  	&   &   & (voxels)  &  (voxels)\\\hline
Spleen	&	0.94	&	0.03	&	1.07	&	2.00	\\
L. Kidney	&	0.93	&	0.04	&	1.06	&	3.00	\\
Gallbladder	&	0.79	&	0.17	&	1.55	&	4.41	\\
Esophagus	&	0.68	&	0.57	&	2.05	&	6.00	\\
Liver	&	0.95	&	0.02	&	1.42	&	4.12	\\
Stomach	&	0.87	&	0.09	&	2.06	&	8.88	\\
Pancreas	&	0.75	&	0.19	&	1.93	&	7.62	\\
Duodenum	&	0.62	&	0.24	&	3.05	&	12.47	\\
 \hline \hline
  \end{tabular}
	\label{tab:segmentation}
\end{table}

The network was trained and evaluated in a 9-fold cross-validation, using the network implementation available in NiftyNet. Briefly, the network, available as \inlinecode{dense$\_$vnet} in NiftyNet, uses a V-shaped structure (with downsampling, upsampling and skip connections) where each downsampling stage is a dense feature stack (i.e. a sequence of convolution blocks where the inputs are concatenated features from all preceding convolution blocks), upsampling is bilinear upsampling and skip connections are convolutions. The loss is a modified Dice loss (with additional hinge losses to mitigate class imbalance) implemented external to NiftyNet and included via a reference in the configuration file. The network was trained for 3000 iterations on whole images (using the \inlinecode{ResizeSampler})  with random affine spatial augmentations.

Segmentation metrics, computed using NiftyNet's \inlinecode{evaluation} action, and aggregated over all folds, are given in Table~\ref{tab:segmentation}. The segmentation with Dice scores closest to the median is shown in Figure~\ref{fig:segmentation}.

%%%

\subsection{Image regression}

Image regression, more specifically, the ability to predict the content of an image given a different imaging modality of the same object, is of paramount importance in real-world clinical workflows. Image reconstruction and quantitative image analysis algorithms commonly require a minimal set of inputs that are often not be available for every patient due to the presence of imaging artefacts, limitations in patient workflow (e.g. long acquisition time), image harmonization, or due to ionising radiation exposure minimization.

An example application of image regression is the process of generating synthetic CT images from MRI data to enable the attenuation correction of PET-MRI images~\citep{burgos2014attenuation}. This regression problem has been historically solved with patch-based or multi-atlas propagation methods, a class of models that are very robust but computationally complex and dependent on image registration. The same process can now be solved using the deep learning architectures similar to the ones used in image segmentation.

As a demonstration of this application, a neural network was trained and evaluated in a 5-fold cross-validation setup using the \inlinecode{net\_regress} application in NiftyNet. Briefly, the network, available as \inlinecode{highresnet} in NiftyNet, uses a stack of residual dilated convolutions with increasingly large dilation factors~\citep{Li:IPMI:2017}. The root mean square error was used as the loss function and implemented as part of NiftyNet as \inlinecode{rmse}. The network was trained for 15000 iterations on patches of size $80\times 80\times 80$, and using the \inlinecode{iSampler}~\citep{Berger:arxiv:2017} for patch selection with random affine spatial augmentations.

\begin{figure}[!t]
  \centering
  \includegraphics[width=0.8\textwidth]{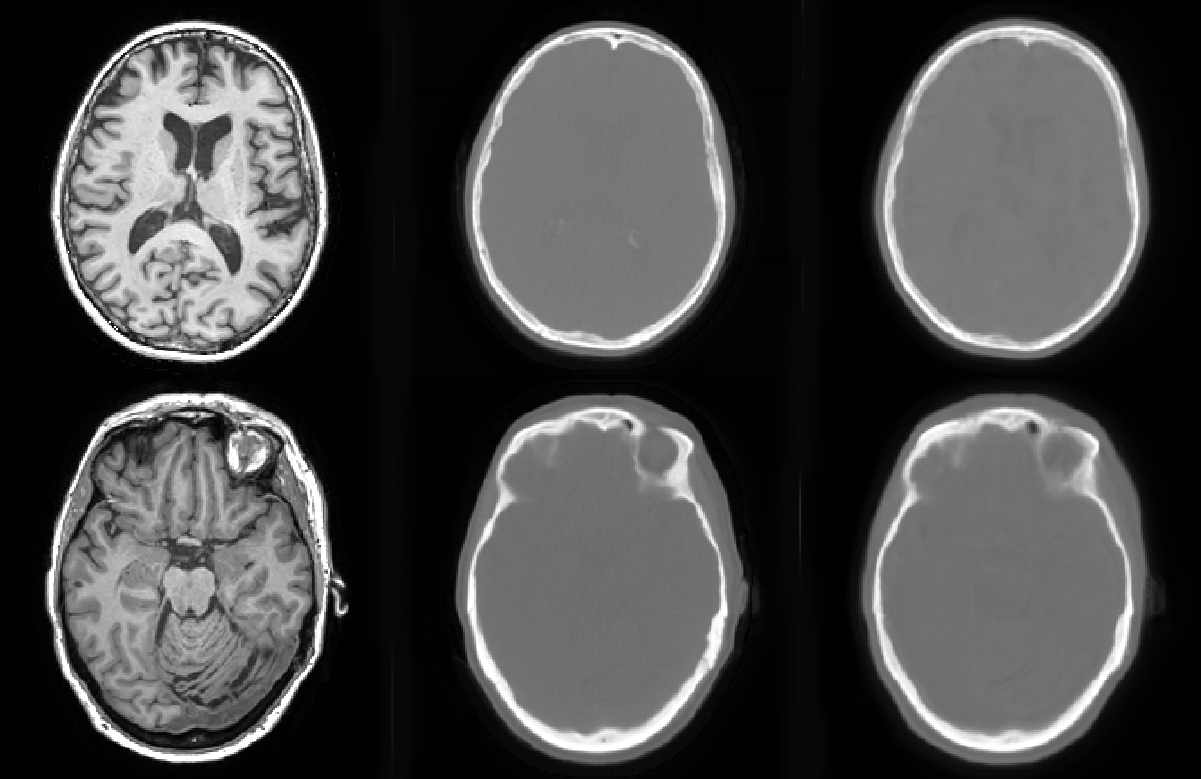}
  \begin{tabularx}{0.8\textwidth}{ X X X }
  	\centering MRI & \centering Ground-truth CT & \centering Synthetic CT
  	\end{tabularx}
  \caption{The input T1 MRI image (left), the ground truth CT (centre) and the NiftyNet regression output (right). }
  \label{fig:regression}
\end{figure}

Regression metrics, computed using NiftyNet's `evaluation` action, and aggregated over all folds, are given in Table~\ref{tab:regression}. The 25th and 75th percentile example result with regards to MAE is shown in Figure~\ref{fig:regression}.

\begin{table}[!b]
	\centering
	\caption{The Mean Absolute Error (MAE) and the Mean Error (ME) between the ground truth and the pseudoCT in Hounsfield units, comparing the NiftyNet method with pCT ~\citep{burgos2014attenuation} and the UTE-based method of the Siemens Biograph mMR.}
\begin{tabular}{cc|cccc}
				&			&NiftyNet		& pCT	& UTE \\
	\hline
\multirow{2}{*}{MAE}	& Average	& 88		& 121	& 203 \\
				& S.D		& 7.5	& 17			& 24 \\
\hline
\multirow{2}{*}{ME}	& Average	& 9.1	& -7.3	& -132 \\
				& S.D.		& 12		& 23			& 34	\\
 \hline
  \end{tabular}
	\label{tab:regression}
\end{table}

\subsection{Ultrasound simulation using generative adversarial networks}
Generating plausible images with specified image content can support training for radiological or image-guided interventional tasks. Conditional GANs have shown promise for generating plausible photographic images~\citep{Mirza:arXiv:2014}. Recent work on spatially-conditioned GANs~\citep{Hu:RAMBO:2017} suggests that conditional GANs could enable software-based simulation in place of costly physical ultrasound phantoms used for training. Here we present an example illustrating a pre-trained ultrasound simulation network that was ported to NiftyNet for inclusion in the NiftyNet model zoo.

The network was originally trained outside of the NiftyNet platform as described in \citep{Hu:RAMBO:2017}. Briefly, a conditional GAN network was trained to generate ultrasound images of specified views of a fetal phantom using 26,000 frames of optically tracked ultrasound. An image can be sampled from the generative model based on a conditioning image (denoting the pixel coordinates in 3D space) and a model parameter (sampled from a 100-D Gaussian distribution).

The network was ported to NiftyNet for inclusion in the model zoo. The network weights were transferred to the NiftyNet network using NiftyNet's \inlinecode{restore$\_$initializer}, adapted from Sonnet~\citep{Reynolds:github:2017}, which enables trained variables to be loaded from networks with different architectures or naming schemes.

\begin{figure}[t]
  \centering
  \includegraphics[width=\textwidth]{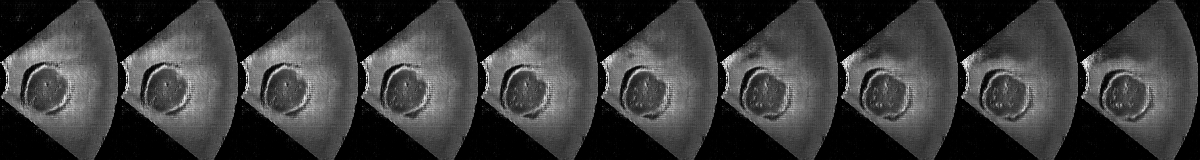}
  \includegraphics[width=\textwidth]{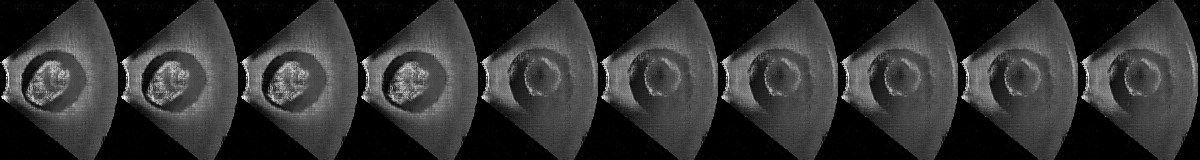}
  \caption{Interpolated images from the generative model space based on linearly interpolated model parameters. The top row shows a smooth variation between different amounts of ultrasound shadow artefacts. The bottom row shows a sharp transition suggesting the presence of mode collapse in the generative model.}
  \label{fig:simulator}
\end{figure}

The network was evaluated multiple times using the \inlinecode{linear$\_$interpolation} inference in NiftyNet, wherein samples are taken from the generative model based on one conditioning image and a sequence of model parameters evenly interpolated between two random samples.  Two illustrative results are shown in Figure~\ref{fig:simulator}. The first shows the same anatomy, but a smooth transition between different levels of ultrasound shadowing artifacts. The second shows a sharp transition in the interpolation, suggesting the presence of mode collapse, a common issue in GANs~\citep{Goodfellow:arXiv:2016}.
%%%
\section{Discussion}
\subsection{Lessons learned}
NiftyNet development was guided by several core principles that impacted the implementation. Maximizing simplicity for simple use cases motivated many implementation choices. We envisioned three categories of users: novice users who are comfortable with running applications, but not with writing new Python code, intermediate users who are comfortable with writing some code, but not with modifying the NiftyNet libraries, and advanced users who are comfortable with modifying the libraries. Support for \inlinecode{pip} installation simplifies NiftyNet for novice and intermediate users. In this context, enabling experimental manipulation of individual pipeline components for intermediate users, and downloadable model zoo entries with modified components for novice users required a modular approach with plugin support for externally defined components. Accordingly, plugins for networks, loss functions and even application logic can be specified by Python \inlinecode{import} paths directly in configuration files without modifying the NiftyNet library. Intermediate users can customize pipeline components by writing classes or functions in Python, and can embed them into model zoo entries for distribution.

Although initially motivated by simplifying variable sharing within networks, NiftyNet's named conceptual blocks also simplified the adaptation of weights from pre-trained models and the TensorBoard-based hierarchical visualization of the computation graphs. The scope of each conceptual blocks maps to a meaningful subgraph of the computation graph and all associated variables, meaning that all weights for a conceptual block can be loaded into a new model with a single scope reference. Furthermore, because these conceptual blocks are constructed hierarchically through the composition of \inlinecode{Layer} objects and scopes, they naturally encode a hierarchical structure for TensorBoard visualization

Supporting machine learning for a wide variety of application types motivated the separation of the \inlinecode{ApplicationDriver} logic that is common to all applications from the \inlinecode{Application} logic that varies between applications. This facilitated the rapid development of new application types. The early inclusion of both image segmentation/regression (mapping from images to images) and image generation (mapping from parameters to images) motivated a flexible specification for the number, type and semantic meaning of inputs and outputs, encapsulated in the \inlinecode{Sampler} and \inlinecode{Aggregator} components.

\subsection{Platform availability}
The NiftyNet platform is available from \url{http://niftynet.io/}. The source code can be accessed from the Git repository\footnote{\url{https://github.com/NifTK/NiftyNet}} or installed as a Python library using \inlinecode{pip install niftynet}. NiftyNet is licensed under an open-source Apache 2.0 license\footnote{\url{https://www.apache.org/licenses/LICENSE-2.0}}. The NiftyNet Consortium welcomes contributions to the platform and seeks inclusion of new community members to the consortium.
%%%
\subsection{Future direction}
The active NiftyNet development roadmap is focused on three key areas: new application types, a larger model zoo and more advanced experimental design. NiftyNet currently supports image segmentation, regression, generation and representation learning applications. Future applications under development include image classification, registration, and enhancement (e.g. super-resolution) as well as pathology detection. The current NiftyNet model zoo contains a small number of models as proof of concept; expanding the model zoo to include state-of-the-art models for common tasks and public challenges (e.g. brain tumor segmentation (BRaTS)~\citep{Menze:TMI:2015,Wang:arXiv:2017a});
and models trained on large data sets for transfer learning will be critical to accelerating research with NiftyNet. Finally, NiftyNet currently supports a simplified machine learning pipeline that trains a single network, but relies on users for data partitioning and model selection (e.g. hyper-parameter tuning). Infrastructure to facilitate more complex experiments, such as built-in support for cross-validation and standardized hyper-parameter tuning will, in the future, reduce the implementation burden on users.
%%%
\section{Summary of contributions and conclusions}
This work presents the open-source NiftyNet platform for deep learning in medical imaging. Our modular implementation of the typical medical imaging machine learning pipeline allows researchers to focus implementation effort on their specific innovations, while leveraging the work of others for the remaining pipeline. The NiftyNet platform provides implementations for data loading, data augmentation, network architectures, loss functions and evaluation metrics that are tailored for the idiosyncracies of medical image analysis and computer-assisted intervention. This infrastructure enables researchers to rapidly develop deep learning solutions for segmentation, regression, image generation and representation learning applications, or extend the platform to new applications.
%%%
\section*{Conflict of interest}
None
\section*{Acknowledgements}
The authors would like to acknowledge all of the contributors to the NiftyNet platform. This work was supported by
the Wellcome/EPSRC [203145Z/16/Z, WT101957, NS/A000027/1];
Wellcome [106882/Z/15/Z, WT103709];
the Department of Health and Wellcome Trust [HICF-T4-275, WT 97914];
EPSRC [EP/M020533/1, EP/K503745/1, EP/L016478/1];
the National Institute for Health Research University College London
Hospitals Biomedical Research Centre (NIHR BRC UCLH/UCL High Impact
Initiative);
Cancer Research UK (CRUK) [C28070/A19985];
the Royal Society [RG160569];
a UCL Overseas Research Scholarship, and a UCL Graduate Research Scholarship.
The authors would like to acknowledge that the work presented here made use of
Emerald, a GPU-accelerated High Performance Computer, made available by the
Science \& Engineering South Consortium operated in partnership with
the STFC Rutherford-Appleton Laboratory;
and hardware donated by NVIDIA.

%%%
\bibliography{JNFull,NiftyNet_2017}

\end{document}